\documentclass[11pt]{article}
\pdfoutput=1
\usepackage[T1]{fontenc}
\usepackage[utf8]{inputenc}
\pdfoutput=1

\usepackage[utf8]{inputenc}
\usepackage{xcolor}
\usepackage{amssymb}
\usepackage{amsfonts}
\usepackage{amsmath}
\usepackage[utf8]{inputenc}
\usepackage{algorithm}
\usepackage{xpatch}
\usepackage{graphicx}
\usepackage{siunitx}
\usepackage{booktabs}
\usepackage{etoolbox}
\usepackage{hyperref}
\usepackage{authblk}
\usepackage{algorithmic}
\usepackage{subfig}
\usepackage[bb=boondox]{mathalfa}

\makeatletter
\usepackage{geometry}
\date{}

\begin{document}
\title{FlowPool: Pooling Graph Representations with Wasserstein Gradient Flows}

\author[]{Effrosyni Simou}
\maketitle

\begin{abstract}
In several machine learning tasks for graph structured data, the graphs under consideration may be composed of a varying number of nodes. Therefore, it is necessary to design pooling methods that aggregate the graph representations of varying size to representations of fixed size which can be used in downstream tasks, such as graph classification. Existing graph pooling methods offer no guarantee with regards to the similarity of a graph representation and its pooled version. In this work, we address this limitation by proposing FlowPool, a pooling method that optimally preserves the statistics of a graph representation to its pooled counterpart by minimizing their Wasserstein distance. This is achieved by performing a Wasserstein gradient flow with respect to the pooled graph representation. Our method relies on a versatile implementation which can take into account the geometry of the representation space through any ground cost and computes the gradient of the Wasserstein distance with automatic differentiation. We propose the differentiation of the Wasserstein flow layer using an implicit differentiation scheme. Therefore, our pooling method is amenable to automatic differentiation and can be integrated in end-to-end deep learning architectures. Further, FlowPool is invariant to permutations and can therefore be combined with permutation equivariant feature extraction layers in GNNs in order to obtain predictions that are independent of the ordering of the nodes. Experimental results demonstrate that our method leads to an increase in performance compared to existing pooling methods when evaluated on graph classification. 
\end{abstract}

\section{Introduction}

In machine learning problems, such as graph classification or graph regression, each datapoint corresponds to a different graph. Examples of such problems arise in the fields of computational biology, drug discovery and social network analysis, among others. For instance, a task of interest is the prediction of whether a protein structure is an enzyme or not. In that case, proteins are represented as graphs with nodes corresponding to amino-acids and edges capturing the spatial proximity of the amino-acids.  Similarly, molecules can be represented as graphs where nodes are atoms and edges the chemical bonds between them. The problem of interest in that case may be the classification of a molecule as active or inactive against cancer cells. 

Such tasks are tackled by models that operate in an inductive setting. In this setting, the goal of graph representation learning algorithms is to use a set of $k$ training graphs $G_1, \ldots G_k$ in order to learn a mapping that can generalize to unseen test graphs $G_{k+1}, \ldots G_{k+l}$. Naturally, the graphs can be of varying size. For instance, in the example of molecule classification discussed above, each molecule may be composed of a different number of atoms. As a result, a relevant problem that arises in Graph Neural Network (GNN) architectures in this context, is to find the optimal way to pool the graph representations of varying size to a representation of fixed size that can be multiplied by the weights of the classifier or regressor that is driving the representation learning process. This operation is commonly referred to as global graph pooling. 

Graph pooling methods proposed fall into two categories: node selection methods and node clustering methods. Node selection methods aim to pool graph representations by selecting the representations of the most important nodes of the graph, according to some criterion. An important disadvantage of these methods is that they discard the information that is captured in the representations of the least important nodes. Node clustering methods propose to pool graph representations by learning node assignment matrices. The motivation of these methods is to learn what constitutes nodes to be similar, so that the representations of similar nodes can be linearly combined, as dictated by the assignment matrices. Node clustering methods don't discard information, but aggregate the representations of  similar nodes. However they still offer no guarantee with regards to the similarity of the graph representation and its pooled version. 

In this work
, we introduce a new type of pooling method, that addresses this limitation. We propose to explicitly preserve the statistics of the graph representation by minimizing the entropy-regularized Wasserstein distance between itself and its pooled version. Our pooling method minimizes this distance by performing a Wasserstein gradient flow with respect to the pooled graph representation. Therefore, we term our proposed pooling method FlowPool. At each step of the flow, the energy of the Wasserstein distance between the graph representation and its pooled counterpart is computed and the pooled representation is moved closer to the original graph representation according to the gradient of that energy. As a result, the pooled graph representation, obtained with FlowPool, combines the representations of the nodes using the optimal couplings that minimize the Wasserstein distance along the steps of the flow.


The structure of this article is as follows. First, in Section \ref{sec:related_work} we review the related work. Further, in Section \ref{sec:preliminaries} we present preliminaries from Optimal Transport theory \cite{villani2008optimal}, \cite{peyre2017computational} that are necessary to the comprehension of the proposed pooling method. After introducing the correspondence of a graph representation to a probability measure in Section \ref{sec:graph2measure}, we propose FlowPool as the minimization of the Wasserstein distance between graph representations in Section \ref{sec:pooling}. In Section \ref{ssec:differentiability} we explain the implementation of FlowPool. In Section \ref{sec:perm_inv} we show that our method is invariant to permutations. In Section \ref{sec:gnn_int} we discuss how we can backpropagate through FlowPool and integrate it in end-to-end GNN architectures.  In Section \ref{sec:pool_experiments} we perform an experimental evaluation of FlowPool. We provide direction for future work in Section \ref{sec:parametrization} and conclude in Section \ref{sec:conclusion}. 

\section{Related Work}\label{sec:related_work}

As mentioned before, existing pooling methods for graphs can be grouped into two categories, namely node selection methods and node clustering methods. We review briefly representative methods from each of these two categories.

Node selection methods pool a graph representation by keeping only the representations of the $K$ most important nodes in the graph. The first node selection method proposed is SortPool \cite{zhang2018end}, which extends the idea of the Weisfeiler-Lehman graph kernels \cite{shervashidze2011weisfeiler} to sort nodes based on their color, to sorting nodes based on their GCN feature vector. Once the nodes have been sorted, only the $K$ most important ones are kept.  Sortpool yields the same representation for isomorphic graphs. A different approach is followed by TopKPool \cite{gao2019graph}. TopKPool proposes to learn from the data a vector of parameters, such that only the $K$ node representations, whose inner product with the trainable vector is maximal, are kept. TopKPool does not explicitly take into account the graph structure. This limitation is addressed by SAGPool \cite{lee_self-attention_2019}. SAGPool employs a trainable graph convolutional layer in order to obtain an attention score for each node in the graph, and keeps only the $K$ nodes that correspond to the highest attention scores. 

Node selection methods lead to loss of information due to the discarded features of the nodes that are not selected. 
Node clustering methods aim to alleviate this problem by finding the optimal way to aggregate the representations of all $N$ nodes in a graph. This is achieved by clustering the $N$ nodes into $M$ clusters in order to obtain an $N\times M$ assignment matrix $S$. As a result, given a graph of $N$ nodes with adjacency matrix $A$ and a $d \times N$ graph representation $Y$, the representation output by node clustering pooling methods is equal to $X=YS$. Further, the assignment matrix $S$ provides a way to obtain the adjacency matrix $A_c$ of the pooled graph as $A_c=S^{\top}AS$. All proposed node clustering methods learn the assignment matrices from the data in order to achieve statistical strength. We provide a summary of existing node clustering pooling methods below.

The first differentiable graph pooling method that proposes to learn assignment matrices is DiffPool \cite{ying2018hierarchical}. DiffPool proposes to parametrize node assignment matrices with a graph convolutional filter followed by a softmax activation. The parameters of the GCN filter are learned from the data during training. Therefore, DiffPool's assignment matrices learn which nodes should be grouped together, based on the features and the graph structure.  Building on DiffPool's concept of parametrizing assignment matrices with GCNs, StructPool \cite{yuan2020structpool} proposes to condition the cluster assignment of each node on the cluster assignments of other nodes. As a result, the graph pooling problem is framed as a structured prediction problem by employing conditional random fields to capture the relationships among the assignments of the nodes. On a different line, HaarPool \cite{Haar}  employs the Haar basis \cite{haar1910theorie}, \cite{hammond2011wavelets} matrix to obtain a mapping of the nodes in the graph to the nodes of the pooled graph. The graph is pooled by keeping the basis vectors that correspond to the low frequencies, thus maintaining the coarse structural information, and discarding the ones that correspond to the high frequencies, thus dismissing fine structural information. In order to leverage the node features as well as the spectral information of the graph structure, MinCutPool \cite{bianchi2020mincutpool} proposes to parametrize node assignment matrices using multi-layer perceptrons and adds a term at the objective function which is a relaxation of the normalized minCUT problem \cite{dantzig2003max}.

Graph pooling is employed in deep networks for graph representation learning in two different contexts; hierarchical pooling and global pooling. Hierarchical pooling aims to generalize the pooling operation, as performed in convolutional networks for regular grids, and aggregate feature information over local patches in order to achieve  locality-preserving representations and invariance to small deformations. Global pooling aims to learn fixed-size representations from graph representations of varying size. In a recent study \cite{mesquita2020rethinking} it is demonstrated that certain graph pooling methods, when employed for hierarchical pooling, do not enhance the graph representation learning process, but rather smooth the features of the nodes in the pooled graph representation. Although in this work we focus on global graph pooling, we mention that our proposed pooling method can also be used for hierarchical pooling and that it could help alleviate the smoothing problem because of the explicit objective of preserving the statistics. In that context, the coarsened graph adjacency matrix $A_c$ could be obtained through the optimal couplings.

Further, we mention that optimal transport-based methods have been recently proposed in order to obtain fixed-size representations. First, the work in \cite{kolouri2020wasserstein} proposes a framework that uses a mapping to obtain a linear approximation of the Wasserstein-2 distance. In order to obtain this mapping, the barycentric projection \cite{ambrosio2008gradient} from a reference to each graph representation is needed. The optimal transport plans, needed for the barycentric projection map calculation, are obtained by solving the Kantorovich problem \cite{kantorovich1942translation} with linear solvers \cite{flamary2021pot}. Once the graph representations of fixed size are obtained, they are used to perform various graph prediction tasks. As this framework is not differentiable, it cannot be integrated in end-to-end architectures for graph representation learning and only non-parametric graph representations are considered. Also, its performance relies heavily on the choice of the values of the representation of the reference, which must be numerically close to those of the graph representations. Second, in \cite{mialon2021trainable} the authors address the problem of finding fixed-size representations of sets of features of varying size and, possibly, in the regime where labeled data are scarse. They introduce an embedding (OTKE) that combines kernel methods \cite{scholkopf2002learning} and optimal transport. They propose to embed the feature representations of a set to a reproducing kernel Hilbert space and subsequently pool the obtained embedding using the optimal transport plan between the kernel embedding and a trainable reference.  Furthermore, the relation of the proposed OTKE to attention mechanisms is discussed and its performance is validated on biological sequence classification and natural language processing tasks. Third, in \cite{naderializadeh2021set} a Euclidean embedding using the generalized sliced Wasserstein distance \cite{kolouri2019generalized} is proposed. It is shown that this embedding can be used to pool representations of sets of varying size in deep neural network architectures.

Finally, we would like to mention that more recently in \cite{sander2021sinkformers} the authors propose Sinkformers
, a variation of the transformer architecture \cite{vaswani2017attention} where the learnable attention matrices are forced to be doubly stochastic using Sinkhorn's algorithm \cite{knight2008sinkhorn}. They consider the case where the attention blocks have tied weights between layers and show theoretically that, in the infinite depth limit, Sinkformers correspond to a Wasserstein gradient flow. 

\section{Preliminaries}\label{sec:preliminaries}
\subsection{The Optimal Transport Problem}
Optimal Transport is a mathematical theory that allows to compare measures in a geometrically meaningful way. Given two discrete probability measures $\mu=\sum_{i=1}^n a_i \delta_{x_i}$ and $ \nu=\sum_{j=1}^m b_j \delta_{y_j}$, the Kantorovich optimal transport problem \cite{kantorovich1942translation} aims to find the optimal probabilistic coupling $P \in \mathbb{R}^{n \times m}_{+}$ between the measures $\mu$ and $\nu$ that minimizes the total cost of transportation:

\begin{equation}\label{eq:kantorovich_objective}
\operatorname*{min}_P \langle C, P \rangle
\end{equation}

while satisfying the constraint:

\begin{equation}\label{eq:mass_preservation}
P \in U(a,b)=\{P \in \mathbb{R}^{n \times m}_{+} | P \mathbb{1}_m=a \text{ and } P^{\top} \mathbb{1}_n=b\}.
\end{equation}

The constraint in Eq. (\ref{eq:mass_preservation}) guarantees that the entirety of the mass of measure $\mu$, namely $a$, is transported to the mass of measure $\nu$, namely $b$. The cost $C$ in Eq. (\ref{eq:kantorovich_objective}) is an $n \times m$ matrix that captures the geometry of the space on which the probability measures are defined, as captured by the pairwise relationships between the points in the support of the measures $\mu$ and $\nu$.

The problem defined in Eq. (\ref{eq:kantorovich_objective}), (\ref{eq:mass_preservation}) is a linear program. Its feasible region is the convex polytope $U(a,b)$ defined by the mass preservation constraints. It can be solved using linear programming solvers such as the simplex algorithm \cite{goldfarb1977practicable}. 

From now on we will denote the Kantorovich optimal transport problem between the measures $\mu=\sum_{i=1}^n a_i \delta_{x_i}$, $ \nu=\sum_{j=1}^m b_j \delta_{y_j}$ as $L_C (a,b)$. This is done in order to highlight the dependency of the problem both on the geometry and the distribution of mass. The geometry is dictated by the supports of the measures $\{x_i\}_{i=1}^n$, $\{y_j\}_{j=1}^m$ and the cost function $C(x,y)$ selected to penalize the transportation of mass from any $x_i$ to any $y_j$. The distribution of mass is dictated by the weight vectors $a$ and $b$. Therefore, from now on:

\begin{equation}\label{eq:Lcab_cost}
L_C (a,b) = \operatorname*{min}_{P \in U(a, b)} \langle P, C \rangle.
\end{equation}

In the specific case where the cost $C$ is the $p$-th power of a metric $D$ on the space $\Omega$, it is shown through the Gluing Lemma (Theorem 7.3 in \cite{villani2008optimal}) that $L_C (a,b)$ can be used to define the $p$-Wasserstein distance between the measures $\mu$, $\nu$ as:

\begin{equation}\label{eq:wass_dist}
W_p(\mu, \nu)= {L_{D^p} (a,b)}^{\frac{1}{p}}.
\end{equation}

Therefore, optimal transport offers a principled way to use the distance between the support of the measures to define a distance between the measures themselves.

\subsection{Entropy Regularization}
Since the Kantorovich problem defined in Eq. (\ref{eq:Lcab_cost}) is a linear program, the complexity of computing $L_C(a,b)$ scales in at least $\mathcal{O}((n+m)nm\log(n+m))$ when comparing discrete measures of $n$ and $m$ points. Also, the problem in Eq. (\ref{eq:Lcab_cost}) attains its minimum at a vertex of the feasible set $U(a,b)$. Therefore, the value $L_C(a,b)$ and the optimal coupling $P^*$ are very susceptible to small changes in the measures $\mu$ and $ \nu$. Both of these issues of high computational complexity and instability to small perturbations are addressed by the entropic regularization of the optimal transport problem \cite{cuturi2013sinkhorn} proposed in the context of machine learning by Cuturi in 2013. 

The entropy regularized optimal transport problem aims to find an optimal probabilistic coupling $P$ between the measures $\mu$ and $\nu$ that minimizes the total cost of transportation and whose negative entropy is as small as specified by a regularization parameter $\epsilon$. Given the entropy $H(P)=-\sum_{i=1}^n\sum_{j=1}^m P_{i,j}\log P_{i,j}$ of the coupling $P$, the regularized problem takes the form:

\begin{equation}\label{eq:entropy_reg}
L_C^{\epsilon}(a, b)=\min_{P \in U(a,b)} \langle C,  P \rangle - \epsilon H(P).
\end{equation} 

The problem in Eq. (\ref{eq:entropy_reg}) can be solved with a fixed point algorithm, by alternating the updates:

\begin{equation}\label{eq:sinkhorn_iterations}
u^{(l+1)}=\frac{a}{Kv^{(l)}} \text{ and } v^{(l+1)}=\frac{b}{K^{\top}u^{(l+1)}}
\end{equation}

where the divisions are applied element-wise and the initialization is a positive vector, for instance $v^{(0)}=\mathbb{1}_m$. The iterations in Eq. (\ref{eq:sinkhorn_iterations}) define Sinkhorn's matrix scaling algorithm \cite{knight2008sinkhorn} on the Gibbs kernel $K=e^{-\frac{C}{\epsilon}}$. Therefore, the solution of the entropy-regularized problem with Sinkhorn's algorithm consists of matrix-vector multiplications and element-wise divisions and its computational complexity is $\mathcal{O}\Big ((n+m)^2 \Big )$ when comparing measures of $n$ and $m$ points. The optimal coupling, as  obtained after $L$ Sinkhorn iterations is equal to:

\begin{equation}\label{eq:optimal_coupling}
P^*=\operatorname{diag}(u^{(L)})K\operatorname{diag}(v^{(L)}).
\end{equation}

By considering dual variables $f,g$, we can obtain the dual formulation of the entropy-regularized transport problem as:

\begin{equation}\label{eq:entropy_reg_dual}
L_C^{\epsilon}(a,b)=\operatorname*{max}_{f \in \mathbb{R}^n,g \in \mathbb{R}^m} \langle f, a \rangle + \langle g, b \rangle - \epsilon \langle e^\frac{f}{\epsilon}, Ke^\frac{g}{\epsilon} \rangle.
\end{equation}

The unconstrained maximization problem in Eq. (\ref{eq:entropy_reg_dual}) can be solved with block coordinate ascent \cite{bertsekas1997nonlinear}.  The potentials $f$ and $g$ are updated alternatively by cancelling the respective gradients in these variables of the objective in Eq. (\ref{eq:entropy_reg_dual}). 

The scaling vectors $u, v$ in Eq. (\ref{eq:sinkhorn_iterations}), (\ref{eq:optimal_coupling}) can be obtained from the dual potentials $f, g$ as:

\begin{equation}\label{eq:scalings_from_duals}
\begin{split}
u=e^{\frac{f}{\epsilon}} \\
v=e^{\frac{g}{\epsilon}} \\
\end{split}
\end{equation} 

and, therefore, it follows that:

\begin{equation}\label{eq:coupling_from_potentials}
P^*= e^{\frac{f^*}{\epsilon}}Ke^{\frac{g^*}{\epsilon}}
\end{equation}
\subsection{Sinkhorn Divergence}
The entropy-regularized optimal transport loss in Eq. (\ref{eq:entropy_reg}), (\ref{eq:entropy_reg_dual}) exhibits an entropic bias. Given a measure $\nu=\sum_{j=1}^N b_j \delta_{y_j}$, the Sinkhorn loss between $\nu$ and itself is equal to $L_{C(Y, Y)}^{\epsilon}(b,b)$, where $C(Y,Y)$ denotes the $N \times N $ cost matrix with the pairwise distances between the points $\{y_j\}_{j=1}^N$. For $\epsilon > 0$, $L_{C(Y, Y)}^{\epsilon}(b,b) \neq 0$ and, as a result, the minimization of $L_{C(X,Y)}^{\epsilon}(a,b)$ with respect to the positions of the measure $\mu$ leads to a biased solution, with the measure $\mu$ having a smaller support than that of the target measure $\nu$. In \cite{pmlr-v89-feydy19a}, it is shown that a principled way to remedy the entropic bias is by the definition of the Sinkhorn divergence:

\begin{equation}\label{eq:sinkhorn_div}
S_{C(X,Y)}^{\epsilon}(a, b) = L_{C(X,Y)}^{\epsilon}(a,b) - \frac{1}{2}L_{C(X,X)}^{\epsilon}(a, a)- \frac{1}{2}L_{C(Y,Y)}^{\epsilon}(b, b).
\end{equation}

The Sinkhorn divergence offers a good approximation of the $p$-th power of the Wasserstein distance.

\section{Graph Representations as Probability Measures}\label{sec:graph2measure}
In order to define a Wasserstein distance between graph representations, we first explain how we correspond the representation of a graph to a probability measure. Given a graph $G$ of $N$ nodes, its representation with $d$ features is a $N \times d$ matrix $Y=[y_1; \ldots y_N] $ where $y_j$ is the $1 \times d$ dimensional representation of the $j$-th node. We propose to describe the representation of the graph $G$ as a probability measure:

\begin{equation}\label{eq:graph_prob_measure}
\nu=\sum_{j=1}^N  b_j \delta_{y_j},
\end{equation}

where $b \in \Sigma_N$ is a histogram. The value of $b_j$ captures the significance of the representation of node $j$. In the case where all nodes are of equal importance we may consider $b_j=\frac{1}{N}, \forall j$. On the contrary, we can consider a non-uniform distribution on the nodes in order to account for some uncertainty in the node representation. For instance, in a social network, we may chose to use higher weights for nodes that correspond to users that have been members of the network for more that one year and whose features decribe them well, and lower weights for users that have just joined the social network.

\begin{figure}[h!]
	\includegraphics[width=\linewidth]{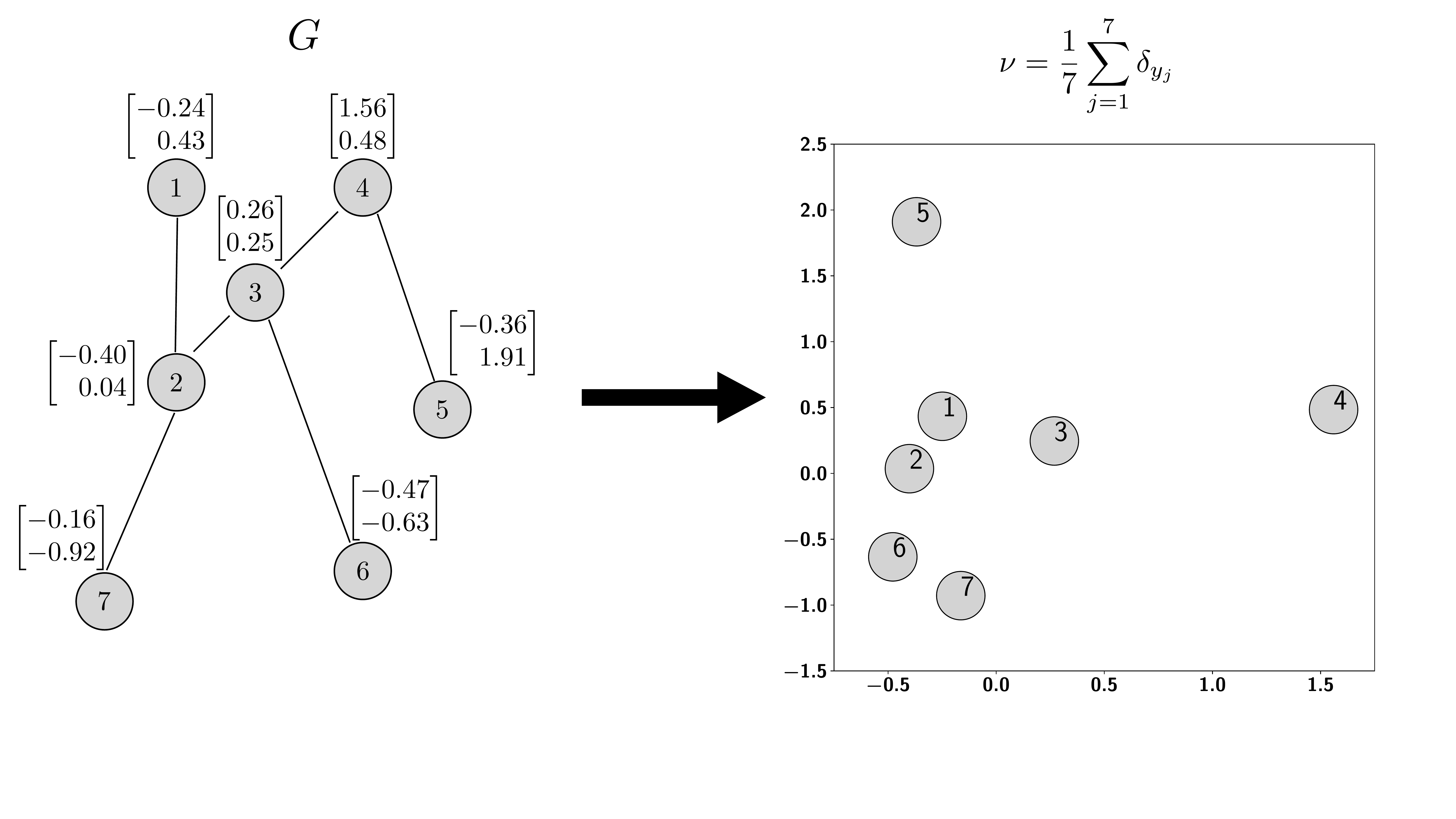}
	\caption{Corresponding a graph $G$ to a probability measure $\nu$. On the left we show a graph $G$ of $N=7$ nodes with its representation $Y$. The $j$-th node is annotated with its representation $y_j=[Y_{j, 1}; Y_{j, 2}]$. For instance, the representation of node $2$ is $y_2=[-0.40; 0.04]$. By assuming that all the nodes in graph $G$ are equally important, we correspond the graph representation to the probability measure $\nu=\frac{1}{7}\sum_{j=1}^7 \delta_{y_j}$. The positions of mass of the measure are determined by the node representations. In the case where the representation $Y$ is obtained by a GNN layer, it implicitly holds information about the graph structure of $G$.}\label{fig:Graph2Measure}
\end{figure}

In Fig. (\ref{fig:Graph2Measure}) we provide an example. We consider a graph $G$ of $N=7$ nodes with representation $Y \in \mathbb{R}^{7 \times 2}$. The $1 \times 2$ dimensional feature representation of node $j$ corresponds to  $y_j=[Y_{j, 1}, Y_{j, 2}]$.  We show on the left of Fig. (\ref{fig:Graph2Measure}) the graph $G$, where each node is annotated with its representation. We correspond to the representation of the graph $G$ the probability measure $\nu=\sum_{j=1}^7 b_j \delta_{y_j}$. Assuming that all nodes have equal importance, the weights of the measure $\nu$ are equal to $b_j=\frac{1}{7}, \forall j$. The positions of mass of the probability measure $\nu$ are determined by the representations of the nodes $y_j$. In the case where the representation $Y$ is obtained by a GNN layer, it implicitly holds information about the graph structure of $G$. On the right of Fig. (\ref{fig:Graph2Measure}) we plot the measure $\nu$ in the 2-dimensional Euclidean space. The considered space does not have to necessarily be Euclidean, but the geometry can be learned. In that case, the cost matrix can be parametrized as $C_{\theta}$, where the parameters $\theta$ are learned from the graph dataset. 

\section{FlowPool}\label{sec:pooling}
The goal of global graph pooling is to transform a representation of a graph of $N$ nodes in a $d$-dimensional feature space to a fixed size representation of $M$ nodes with the same feature dimension $d$.
Let $\nu=\sum_{j=1}^N b_j \delta_{y_j}$ and $\mu=\sum_{i=1}^M a_i \delta_{x_i}$ be the probability measures that correspond to the graph representation $Y \in \mathbb{R}^{N \times d}$ and its pooled counterpart $X \in \mathbb{R}^{M \times d}$. FlowPool computes the pooled representation $X$ by solving:

\begin{equation}\label{eq:reg_flow}
\operatorname*{argmin}_X S_{C(X, Y)}^{\epsilon}(a, b),
\end{equation}

where $S^{\epsilon}_{C(X,Y)}(a,b)$ is the Sinkhorn divergence defined in Eq.(\ref{eq:sinkhorn_div}) for a cost $C=D^p$, with $D$ a distance metric on the graph representation space $\Omega=\mathbb{R}^d$. We denote the cost of the mass transportation as $C(X, Y)$ in order to highlight the dependency of the cost $C$ on $X, Y$. From now on, we assume that the nodes are of equal importance in all cases, so that $a=\frac{1}{M}\mathbb{1}_M$ and $b=\frac{1}{N}\mathbb{1}_N$.  

\begin{figure}[h!]
	\centering
	\includegraphics[scale=0.5]{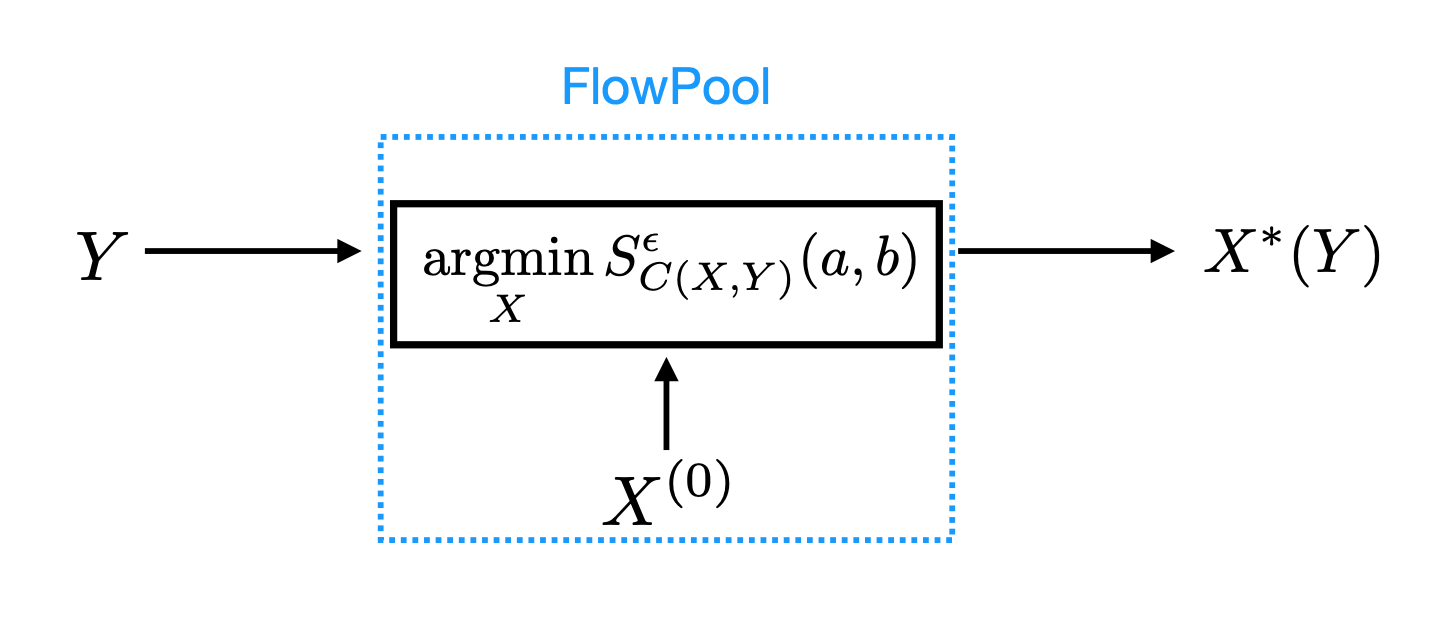}
	\caption{The FlowPool method for pooling graph representations. The input is a graph representation $Y \in \mathbb{R}^{N \times d}$, where $N$ can admit any value. The output $X^{*}$ is a $\mathbb{R}^{M \times d}$ representation, with $M$ fixed, such that $S^{\epsilon}_{C(X^{*}, Y)}(a,b)$ is minimal.}\label{fig:FlowPool}
\end{figure}

The objective function in Eq. (\ref{eq:reg_flow}) is differentiable with respect to $X$. Therefore, we can compute the pooled graph representation by solving the problem in Eq. (\ref{eq:reg_flow}) with a gradient-based optimization method. Thus, our proposed pooling method is cast to a gradient flow. By denoting as $X^{(0)}$ the initialization for the pooled graph representation and as $X^{*}$ the optimal solution of Eq. (\ref{eq:reg_flow}), FlowPool can be thought of as demonstrated in Fig. (\ref{fig:FlowPool}). Specifically, given an initialization $X^{(0)} \in \mathbb{R}^{M \times d}$, the FlowPool layer summarizes any graph representation $Y \in \mathbb{R}^{N \times d}$, by performing a gradient-based method that minimizes the energy $S^{\epsilon}_{C(X, Y)}(a,b)$. 
The returned pooled representation $X^{*}$ is the $M \times d$ representation that has minimal Wasserstein distance from the $N \times d$ input representation $Y$. 

\section{Implementation of FlowPool}\label{ssec:differentiability}
At each step of the gradient-based optimization of $X$, we need to compute the gradient with respect to $X$ of the energy $S_{C(X, Y)}^{\epsilon}(a, b)$ for the current $X$. One possible way to compute this gradient is to unroll the iterations of the Sinkhorn algorithm. A more efficient approach to compute $\nabla_X S_{C(X, Y)}^{\epsilon}(a, b)$ is by using the implicit function theorem \cite{krantz2012implicit}. The implicit differentiation is particularly more efficient in terms of memory requirements as not all intermediate calculations of the Sinkhorn iterations need to be stored in memory for the backpropagation.
In this work we use the \texttt{OTT} toolbox \cite{ott-jax}, which provides implementations for the automatic differentiation of the entropy-regularized optimal transport problem both with implicit differentiation \cite{cuturi2020supervised} and with unrolling \cite{adams2011ranking}. 

We show in Algorithm \ref{algo:general_FlowPool}, the implementation of FlowPool as a Wasserstein gradient flow on the pooled representation. In line 3 we compute the optimal potentials $f^*, g^*$ of $L_{C(X,Y)}^{\epsilon}(a,b)$, $f_{xx}^*, g_{xx}^*$ of $L_{C(X,X)}^{\epsilon}(a,a)$ and $f_{yy}^*, g_{yy}^*$ of $L_{C(Y,Y)}^{\epsilon}(b,b)$ using block coordinate ascent, as described in Section \ref{sec:preliminaries}. We use automatic differentiation in line 5 to obtain the gradient $\nabla_{X} S_{C(X^{(l)}, Y)}^{\epsilon}(a, b)$ and update the pooled representation using this gradient in line 7. The geometry is captured by the cost function $C$. We denote with $\tau$ the step size used for the gradient flow. 

\begin{algorithm}[H]
	\caption{FlowPool}
	\begin{algorithmic}[1]
		\renewcommand{\algorithmicrequire}{\textbf{Input:}}
		\REQUIRE $ Y; X^{(0)} $
		\WHILE {not converged}
		\STATE Solve Sinkhorn 
		\STATE $f^*, g^*, f_{xx}^*, g_{xx}^*, f_{yy}^*, g_{yy}^* = \text{Sinkhorn} \big (a, b, X^{(l)}, Y, C, \epsilon \big )$
		\STATE Compute gradient of Sinkhorn divergence w.r.t. pooled representation
		\STATE $\nabla_{X} S_{C(X^{(l)}, Y)}^{\epsilon}(a, b)=\text{sinkhorn\_diff} \big (f^*, g^*, f_{xx}^*, g_{xx}^*, f_{yy}^*, g_{yy}^*, a, b, X^{(l)}, Y, C, \epsilon \big )$
		\STATE Update pooled graph representation
		\STATE $X^{(l+1)}=\text{update\_step} \Big (X^{(l)},  \nabla_{X} S_{C(X^{(l)}, Y)}^{\epsilon}(a,b), \tau \Big )$
		\ENDWHILE
		\RETURN $X^{*}$
	\end{algorithmic}\label{algo:general_FlowPool}
\end{algorithm}

\section{Permutation Invariance}\label{sec:perm_inv}
We now show that the proposed pooling method is permutation invariant. This is a particularly relevant property when performing global pooling of graph representations. If FlowPool is preceded by permutation equivariant message passing layers \cite{maron2018invariant}, the pooled graph representation is permutation invariant. This means that the predictions made for a given graph are independent of the ordering of its nodes.

\subsection{Computation of $\nabla_X S_{C(X,Y)}^\epsilon (a, b)$}

Although the gradient $\nabla_X S_{C(X,Y)}^\epsilon (a, b)$ is computed in an automatic way in FlowPool, we demonstrate now how it can be computed analytically in order to show that FlowPool is invariant to permutations. It holds that:

\begin{equation}\label{eq:sinkhorn_div_gradient}
\nabla_X S_{C(X,Y)}^\epsilon (a, b) = \nabla_X L_{C(X,Y)}^\epsilon (a, b) - \frac{1}{2} \nabla_X L_{C(X,X)}^\epsilon (a, a).
\end{equation}

In the case of FlowPool we can consider each term in Eq. (\ref{eq:sinkhorn_div}) to be a parametrized optimization problem with parameter the cost $C$. Therefore, we can express $L_C^\epsilon (a, b)$ as: 
\begin{equation}\label{eq:dual_reg_ot}
L_C^{\epsilon}(a,b)=\operatorname*{max}_{f \in \mathbb{R}^n,g \in \mathbb{R}^m} Q(f(C) ,g(C), C)
\end{equation}

where:

\begin{equation}\label{eq:value_function}
Q(f(C) ,g(C), C) = \langle f(C), a \rangle + \langle g(C), b \rangle - \epsilon \langle e^\frac{f(C)}{\epsilon}, e^{-\frac{C}{\epsilon}}e^\frac{g(C)}{\epsilon} \rangle
\end{equation}

Due to the envelope theorem \cite{afriat1971theory}, \cite{takayama1985mathematical}, we can consider that changes in the optimization variables $f^*, g^*$ do not contribute to the changes in the objective function $Q(f(C) ,g(C), C)$, when $C$ is perturbed. Therefore, it follows from Eq. (\ref{eq:value_function}), (\ref{eq:coupling_from_potentials}) that the gradient of $Q$ with respect to $C_{i,j}$ is equal to:

\begin{equation}
\frac{\partial Q(f^*(C) ,g^*(C), C)}{\partial C_{ij}}
= - \epsilon e^{\frac{f^*_i (C)}{\epsilon}} \frac{\partial  e^{-\frac{C_{ij}}{\epsilon}}}{\partial C_{ij}} e^{\frac{g_j^*(C)}{\epsilon}}=
e^{\frac{f^*_i (C)}{\epsilon}} e^{-\frac{C_{ij}}{\epsilon}} e^{\frac{g_j^*(C)}{\epsilon}}= P_{ij}^*.
\end{equation}

As a result, from the chain rule, we obtain that:

\begin{equation}\label{eq:final_gradient}
\nabla_X L_C^{\epsilon}(a,b)= [\partial_X  C(X,Y)]^{\top}  P^* 
\end{equation}

From Eq. (\ref{eq:sinkhorn_div_gradient}), (\ref{eq:final_gradient}) we obtain that:

\begin{equation}
\nabla_X S_{C(X,Y)}^\epsilon (a, b) = [\partial_X  C(X,Y)]^{\top}  P^*  - \frac{1}{2} [\partial_X  C(X,X)]^{\top}  P_{xx}^* .
\end{equation}

\subsection{Proof of Permutation Invariance}

We consider a graph representation $Y_1 \in \mathbb{R}^{N \times d}$ and its row-wise, or equivalently node-wise, permutation according to an $N \times N$ permutation matrix $\mathcal{P}_f$:

\begin{equation}\label{eq:feature_perm}
Y_2 = \mathcal{P}_f Y_1.
\end{equation}

We will show that $\text{FlowPool}(Y_1)=\text{FlowPool}(Y_2)$.

\subsubsection{OT cost}
Let $X^{(0)} \in \mathbb{R}^{M \times d}$ be the initialization of the pooled graph representation and $C_1$, $C_2$ be the $M \times N$ cost matrices that capture the pairwise squared Euclidean distances from $X^{(0)}$ to $Y_1$ and to $Y_2$, respectively. From Eq.(\ref{eq:feature_perm}) we obtain:

\begin{equation}\label{eq:cost_perm}
C_2 = \mathcal{P}_f C_1.
\end{equation}

\subsubsection{OT coupling}
Because of the entropic regularization, the problem in Eq. (\ref{eq:entropy_reg}) is a strictly convex minimization problem. Given uniform weight distributions, $a=\frac{1}{M}\mathbb{1}_M$ and $b=\frac{1}{N}\mathbb{1}_N$, the only variable controlling the unique solution $P^*$ is the cost for the transportation. Therefore, due to the relationship between the costs $C_1, C_2$ in Eq.(\ref{eq:cost_perm}), it holds that:

\begin{equation}\label{eq:coupling_equality}
P^*_2=\mathcal{P}_f P^*_1. 
\end{equation}

\subsubsection{Gradient}\label{sssec:gradient}
Due to Eq. (\ref{eq:cost_perm}), (\ref{eq:coupling_equality}), and the expression for the gradient of $L_C^{\epsilon}(a,b)$ with respect to $X$ in Eq. (\ref{eq:final_gradient}), it follows that the gradient $\nabla_X L_{C(X, Y)}^{\epsilon}(a,b)$ will be the same for both the permuted and the non-permuted case. Further, permutations in $Y$ don't affect the term $\nabla_X L_{C(X,X)}^\epsilon (a, a)$ and therefore:

\begin{equation}
\nabla_X S_{C(X, Y_1)}^{\epsilon} (a, b)= \nabla_X S_{C(X, Y_2)}^{\epsilon} (a, b)
\end{equation}

\subsubsection{Update Step}
As a result, if the initialization $X^{(0)}$ is common in both cases, then $X_1^{(1)} =X_2^{(1)}$. Similarly, we can show that this is true for all iterations of the flow. Therefore, it holds that $X_1^{*} =X_2^{*} \Leftrightarrow \text{FlowPool}(Y_1)=\text{FlowPool}(Y_2)$. The assumption that $X^{(0)}$ is common for both cases is a reasonable one. In fact, the common initialization $X^{(0)}$ serves as a point of reference in the representation space and guarantees that the proposed pooling method performs well. 

\section{Integrating FlowPool in Graph Neural Network Architectures}\label{sec:gnn_int}
\begin{figure}[h!]
	\centering
	\includegraphics[width=\linewidth]{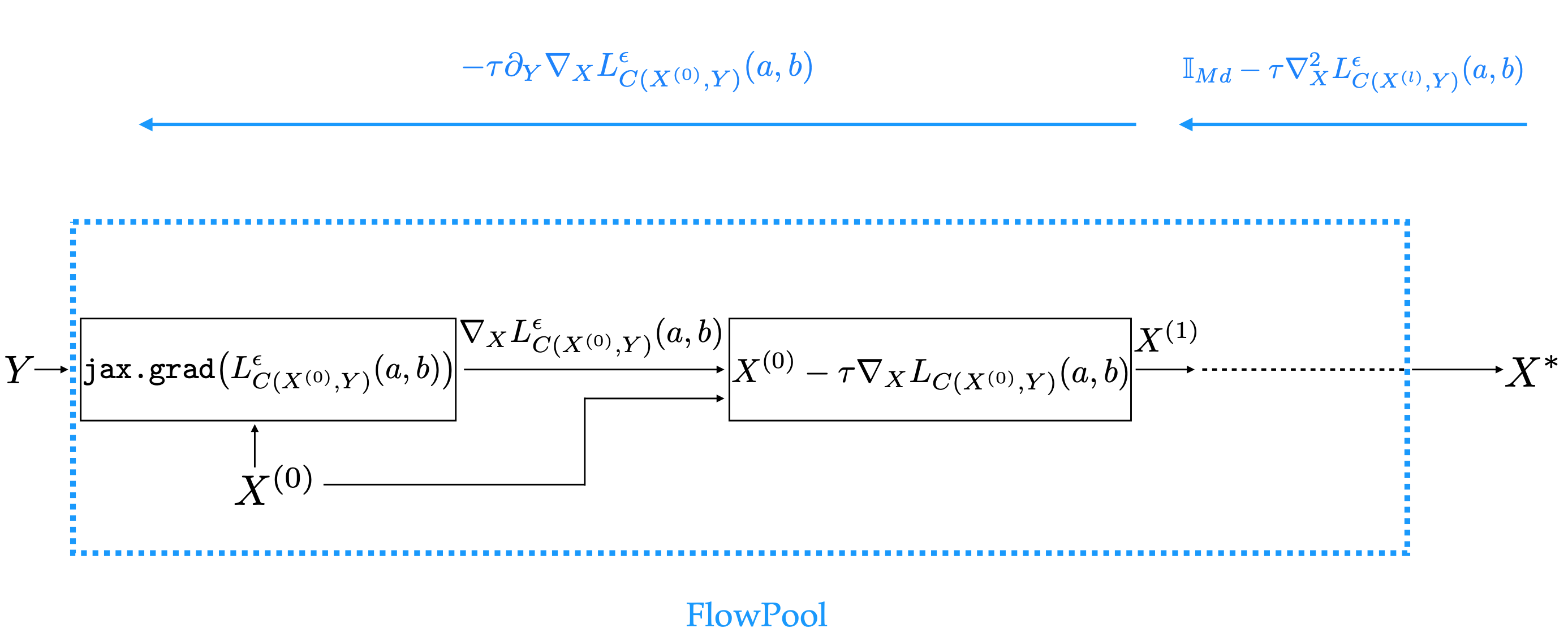}
	\caption{Implementation of FlowPool using JAX. The gradient $\nabla_{X} L_{C(X, Y)}^{\epsilon}(a, b)$ needed for the gradient flow is obtained with JAX's function \texttt{jax.grad}. The pooled graph representation is updated according to this gradient. During the backpropagation the gradient function obtained with \texttt{jax.grad} is re-derived in order to obtain $\partial_Y \nabla_{X} L_{C(X, Y)}^{\epsilon}(a, b)$ and $ \nabla^2_{X} L_{C(X, Y)}^{\epsilon}(a, b)$.} \label{fig:backprop}
\end{figure}

In order to integrate FlowPool in end-to-end deep learning architectures for graphs an element that needs to be considered is the differentiation of the output of FlowPool with respect to its input. This differentiation is key in the general setting, where the input representation $Y$ can be the output of one or more layers with trainable weights and we need to backpropagate through FlowPool in order to learn these weights. In order to backpropagate through FlowPool we need to compute the Jacobian of the output $X^{*}$ with respect to the input $Y$. In this section we outline how this computation is handled. We depict a block-scheme of the computations in FlowPool in Fig. (\ref{fig:backprop}) for the case where the update step of the flow corresponds to that of gradient descent. Further, we assume here for simplicity that FlowPool only minimizes the first term $L_{C(X,Y)}^\epsilon (a,b)$ of the Eq. (\ref{eq:sinkhorn_div}).

In order to compute the Jacobian $\partial_Y X^*$, the Jacobian of the first step of the flow $\partial_Y X^{(1)}$ as well as the Jacobians at the intermediate steps of the flow $\partial_{X^{(l)}}X^{(l+1)}$ for $l=1, \ldots, L-1$ are needed. In the case of a gradient descent update these are respectively:

\begin{equation}\label{eq:jacobian1}
\begin{split}
\partial_Y X^{(1)} &=  \partial_Y \Big ( X^{(0)} - \tau \nabla_X L_{C(X^{(0)}, Y)}^{\epsilon} (a, b) \Big )\\
&=-\tau \partial_Y \nabla_X L_{C(X^{(0)}, Y)}^\epsilon (a,b)
\end{split}
\end{equation}

and


\begin{equation}\label{eq:jacobian2}
\begin{split}
\partial_{X^{(l)}}X^{(l+1)} &= \partial_{X^{(l)}} \Big ( X^{(l)} - \tau \nabla_X L_{C(X^{(l)}, Y)}^{\epsilon} (a, b) \Big )\\
&= \partial_{X^{(l)}} X^{(l)} - \tau  \partial_{X^{(l)}} \nabla_X L_{C(X^{(l)}, Y)}^{\epsilon} (a, b) \\
&= \mathbb{I}_{Md} - \tau \nabla_X^2 L_{C(X^{(l)}, Y)}^{\epsilon} (a, b),
\end{split}
\end{equation}

where $\nabla^2$ stands for the Hessian matrix with second-order derivatives.

\begin{figure}[!h]
    \centering
    \subfloat[]{{\includegraphics[width=0.45 \linewidth]{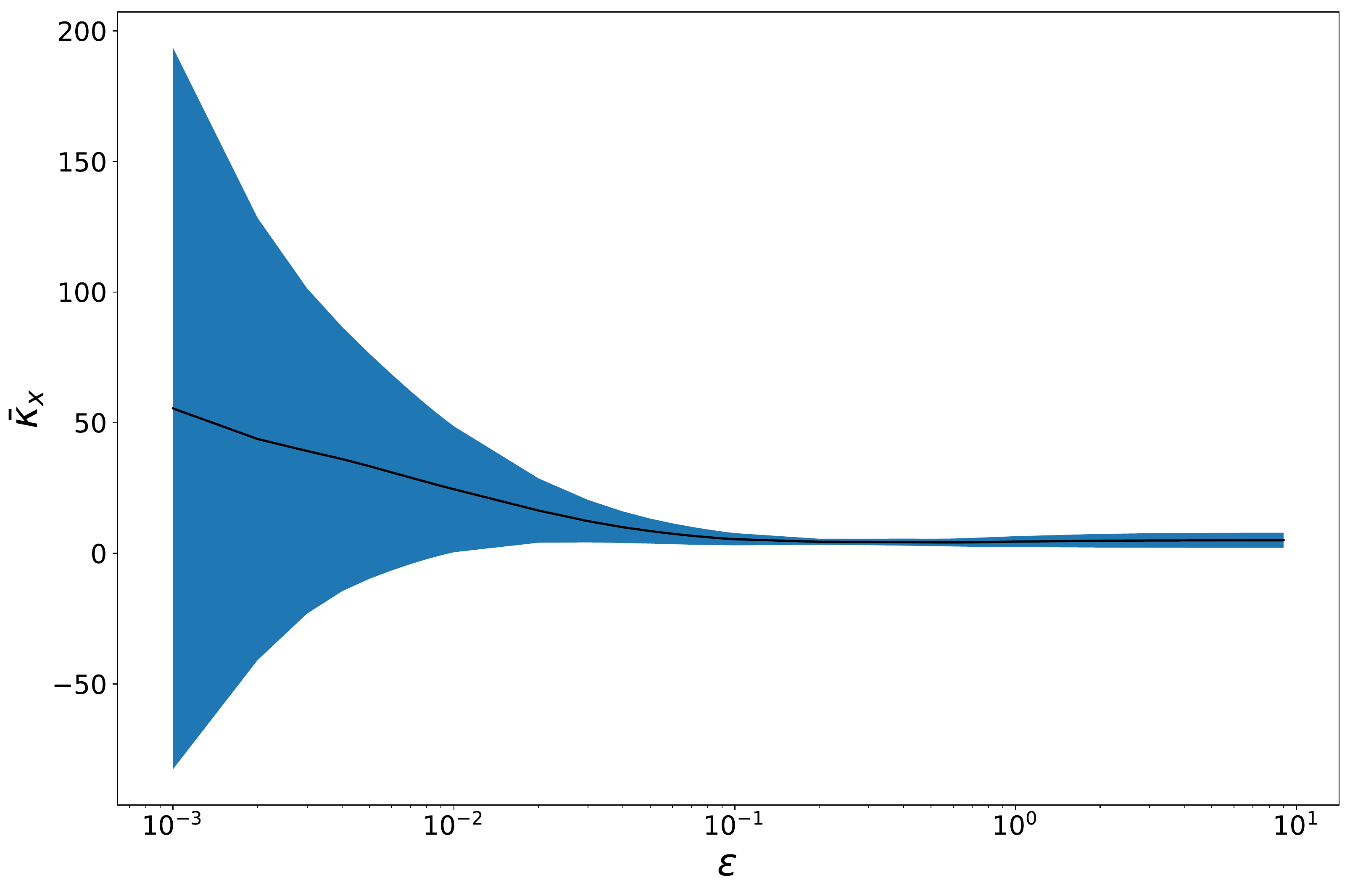} }}%
    \qquad
    \subfloat[]{{\includegraphics[width=0.45 \linewidth]{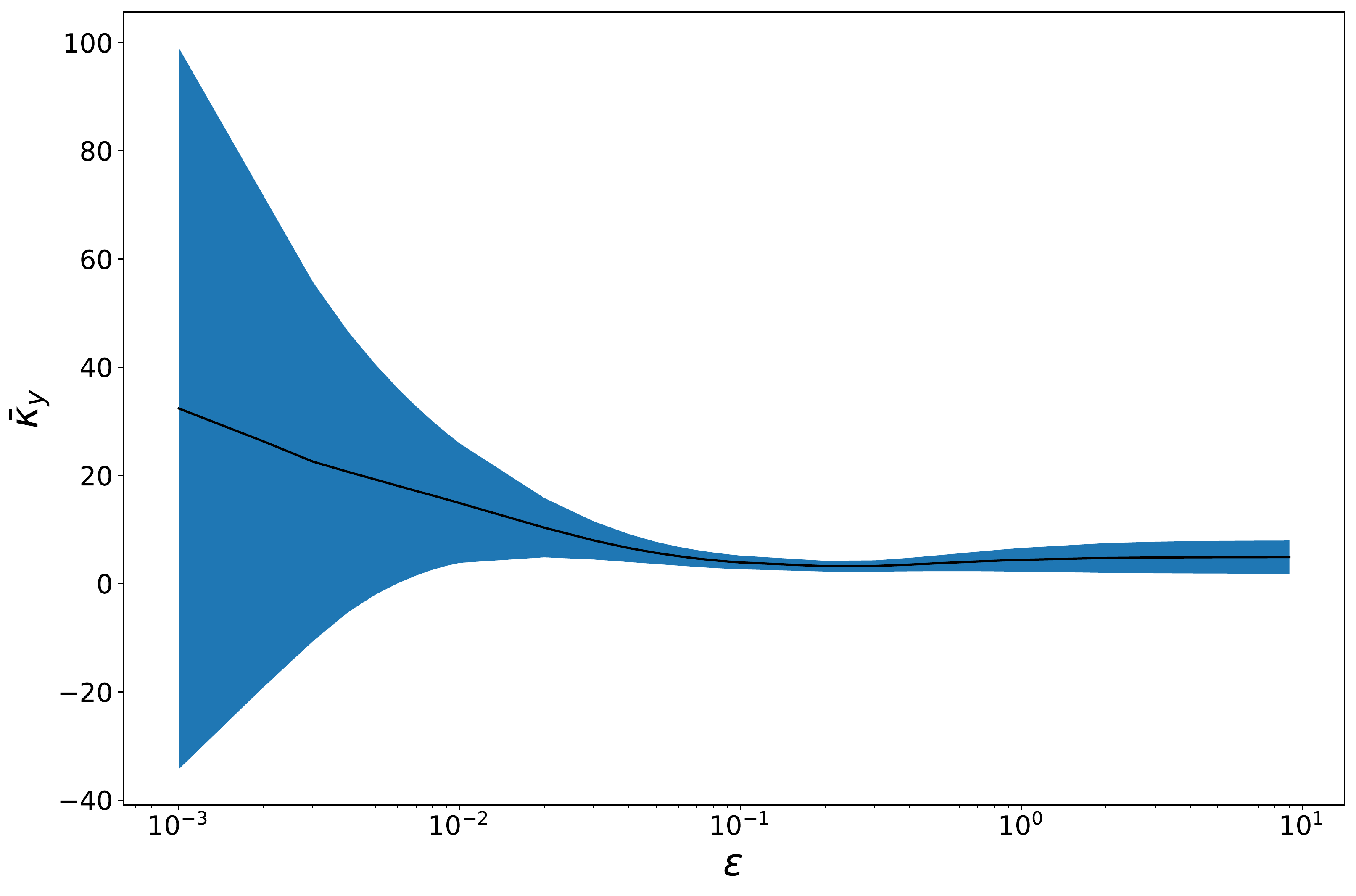} }}%
    \caption{Evolution of the condition numbers $\kappa_x$, $\kappa_y$ of the gradient function $\nabla_X L_{C(X, Y)}^\epsilon (a,b)$ with respect to the entropic regularization parameter $\epsilon$. The black line shows the mean and the blue shaded area the standard deviation. Larger values of $\epsilon$ lead to better conditioning.}%
    \label{fig:condition_numbers}%
\end{figure}

Therefore, we must be able to differentiate the gradient function $\nabla_X L_{C(X^{(0)}, Y)}^\epsilon (a,b)$ with respect to $Y$ and $X$. The Sinkhorn loss in Eq. (\ref{eq:entropy_reg}) is smooth with respect to $C(X, Y)$ \cite{peyre2019computational}. Therefore, the gradient of the Sinkhorn loss is a continuous function. If it is also differentiable, we can measure how much the output value of $\nabla_X L_{C(X, Y)}^\epsilon (a,b)$ can change for a small change in the input arguments $X$ or $Y$ by computing the condition numbers as:

\begin{equation}
\kappa_x = \frac{\|\partial_X \nabla_X L_{C(X, Y)}^\epsilon (a,b)\|}{\frac{\| \nabla_X L_{C(X, Y)}^\epsilon (a,b)\|}{\|X\|}}=\frac{\sigma_{max}(\nabla^2_X  L_{C(X, Y)}^\epsilon (a,b))}{\frac{\| \nabla_X L_{C(X, Y)}^\epsilon (a,b)\|}{\|X\|}}
\end{equation}

and

\begin{equation}
\kappa_y = \frac{\|\partial_Y \nabla_X L_{C(X, Y)}^\epsilon (a,b)\|}{\frac{\| \nabla_X L_{C(X, Y)}^\epsilon (a,b)\|}{\|Y\|}}=\frac{\sigma_{max}(\partial_Y \nabla_X L_{C(X, Y)}^\epsilon (a,b))}{\frac{\| \nabla_X L_{C(X, Y)}^\epsilon (a,b)\|}{\|Y\|}}
\end{equation}

where $\sigma_{max}$ is the maximal singular value. We consider values of $\epsilon$ in $[0.001, 10.0]$ and compute the condition numbers $\kappa_x, \kappa_y $ for 70 randomly generated pointclouds. We show in Fig. (\ref{fig:condition_numbers}) the mean and the standard deviation for $\kappa_x$ and $\kappa_y$ averaged over the 70 pointclouds. It can be seen that for small values of $\epsilon$, the condition numbers $\kappa_x, \kappa_y$ are large and that they have a large variance. As $\epsilon$ increases, the condition numbers and their variance decrease and, eventually, $\kappa_x$ and $\kappa_y$ tend to the value of 1. Therefore, it follows that for small values of $\epsilon$, the derivatives of the function $\nabla_X L_{C(X, Y)}^\epsilon (a,b)$ can admit very large values, whereas larger values of $\epsilon$ lead to better conditioning. As a result, the gradient of the Sinkhorn loss has well-defined derivatives as long as $\epsilon$ is sufficiently large.

\begin{figure}[!h]
\centering
	\includegraphics[width=0.8\linewidth]{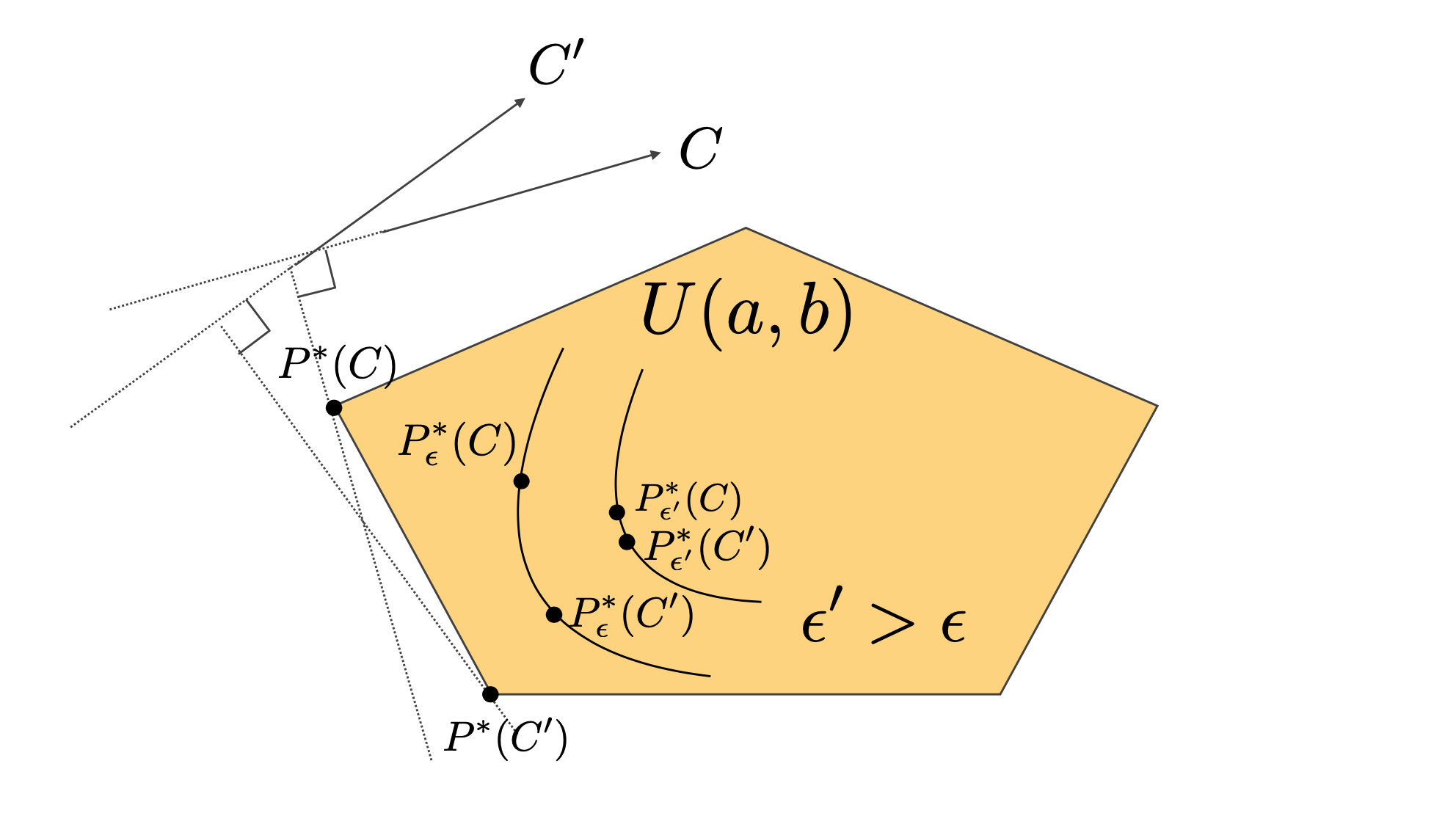}
	\caption{Effect of the entropic regularization parameter $\epsilon$ on the change of the optimal coupling to changes in the cost. For a low value of $\epsilon$, small changes in the cost, lead to a large change in the optimal coupling. For a larger regularization $\epsilon^\prime$, small changes in the cost, lead to smaller changes in the optimal solution $P_{\epsilon^\prime}^*$.}
	\label{fig:linear_vs_sinkhorn}
\end{figure}  

\begin{figure}[!h]
\centering
	\includegraphics[width=0.5\linewidth]{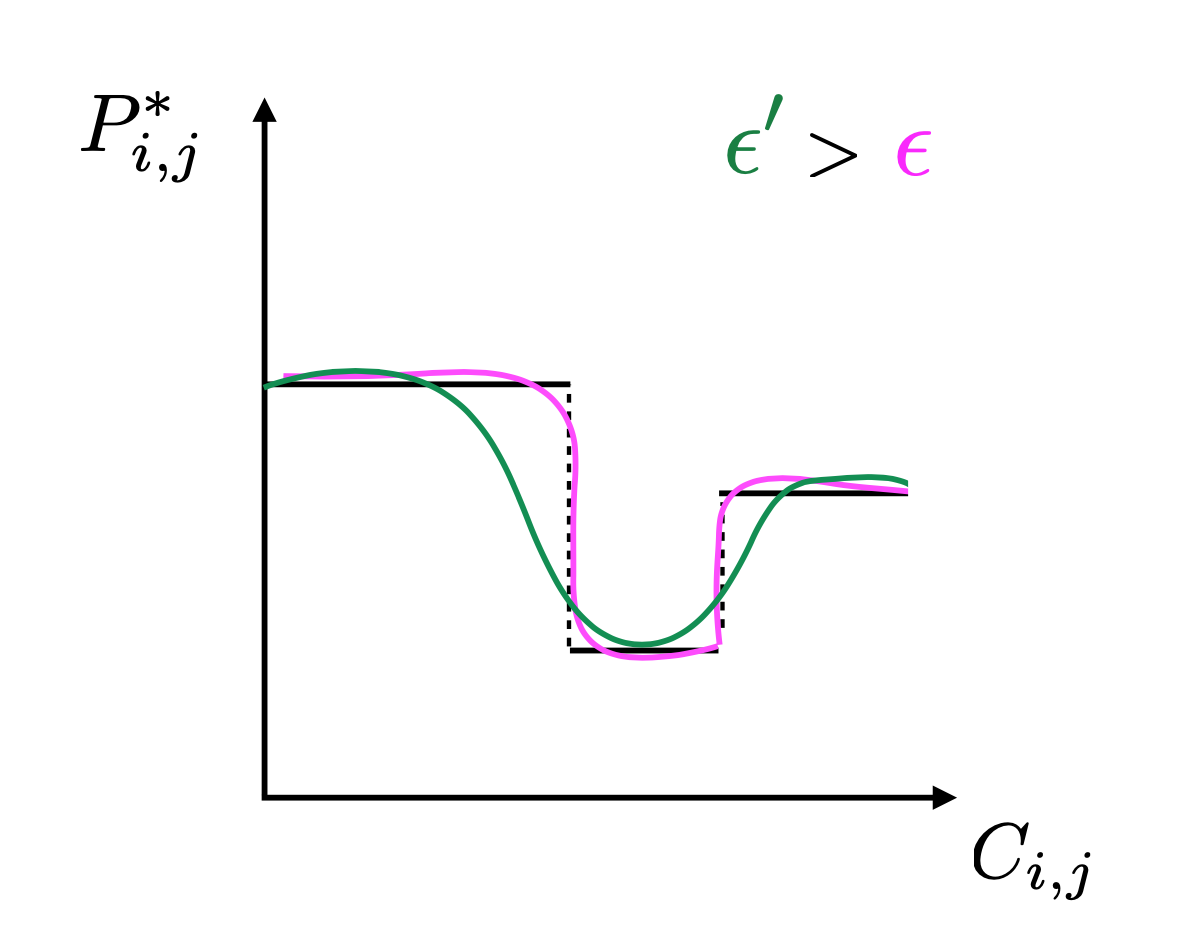}
	\caption{Plot of $P_{i,j}^*$ as a function of $C_{i,j}$. In the case of no regularization (shown with the black plot), changes in $C$ lead to no change in $P^*$, until the perturbation in $C$ is significant enough for the solution of the linear program to move to another vertex. This results to an abrupt change and a step-like function. In the case of a small regularization parameter $\epsilon$ (pink plot), we obtain a continuous function that is not differentiable at the values of $C_{ij}$ that result to an optimal coupling that is close to another vertex. For a larger regularization parameter $\epsilon^\prime$, the obtained function is smooth enough (green plot) so that the gradient function of Sinkhorn has well defined derivatives. }
	\label{fig:pij_wrt_cij}
\end{figure}

In order to explain why this is the case, we show in Fig. (\ref{fig:linear_vs_sinkhorn}) how changes in the cost $C$, due either to changes in $X$ or $Y$, affect changes in the optimal solution $P^*$ of the problem in Eq. (\ref{eq:entropy_reg}). The unregularized OT problem in Eq. (\ref{eq:kantorovich_objective}) is a linear program described by the polytope of mass preservation constraints $U(a,b)$ and the plane $C(X,Y)$. The optimal solution $P^*(C)$ to that problem is the vertex of the polytope $U(a,b)$ that minimizes the dot product with the cost $C$. The solution of the respective entropy regularized OT problem is within $U(a,b)$ and the proximity of it to $P^*(C)$ depends on the strength of the regularization parameter $\epsilon$. We illustrate this in Fig. (\ref{fig:linear_vs_sinkhorn}), where we show the solution $P^*(C)$ of the unregularized OT problem for cost $C$ and the solutions $P_{\epsilon}^*(C), P_{\epsilon^\prime}^*(C)$ of the regularized problem for parameters $\epsilon^\prime > \epsilon$. We can consider a small perturbation in $X$ or $Y$ that results in a new cost $C^\prime$. In that case, the optimal solution $P^*(C^\prime)$  is another vertex of the polytope $U(a,b)$. As, a result, small changes in $C$ result to very large changes in the solution of the unregularized OT problem. For the regularized problem, when $\epsilon$ is very small, similar behaviour is seen with that of the unregularized problem. Therefore, for a small $\epsilon$, a minimal perturbation from $C$ to $C^\prime$ results to large change from $P_\epsilon^*(C)$ to $P_\epsilon^*(C^\prime)$. However, in the case of stronger regularization $\epsilon^\prime > \epsilon$, the change from $P_{\epsilon^\prime}^*(C)$ to $P_{\epsilon^\prime}^*(C^\prime)$ is smaller, as shown in Fig. (\ref{fig:linear_vs_sinkhorn}). It can be seen from Eq. (\ref{eq:entropy_reg}) that the gradient of $L_{C(X,Y)}^\epsilon (a, b)$ with respect to $C$ depends partly on the Jacobian of $P_{\epsilon}^*$ with respect to $C$. Small values of $\epsilon$ result in values in $\partial_C P_\epsilon^*$ that are either $0$ or infinite, due to the vertical tangent line, as shown in Fig. (\ref{fig:pij_wrt_cij}). As a result, for small values of $\epsilon$, the gradient is not a well conditioned function. On the contrary, larger values of $\epsilon$, lead to a smoother dependence of $P_\epsilon^*$ on $C$, as shown in Fig. (\ref{fig:pij_wrt_cij}), and, therefore, to a well conditioned gradient function $\nabla_X L_{C(X, Y)}^\epsilon (a,b)$. Therefore, for sufficiently large values of $\epsilon$ we can compute the Jacobians in Eq. (\ref{eq:jacobian1}), (\ref{eq:jacobian2}) and differentiate through FlowPool. The automatic computation of higher order derivatives of $L_{C(X,Y)}^\epsilon (a, b)$ is possible with \texttt{JAX} and \texttt{OTT}.

\subsection{Unrolling the iterations of the Wasserstein Gradient Flow}
In order to backpropagate through FlowPool, the most straightforward way is to differentiate through all of the iterations of the Wasserstein gradient flow. In that case, if we assume that the gradient flow has converged after $L$ iterations, the Jacobian of the output of FlowPool with respect to its input is $\partial_Y X^{(L)}$ and it is equal to:

\begin{equation}\label{eq:mult_jacobians}
\partial_Y X^{(L)} = \partial_{X^{(L-1)}} X^{(L)} \cdots  \partial_{X^{(1)}} X^{(2)} \partial_{Y} X^{(1)} .
\end{equation}

From Eq. (\ref{eq:jacobian1}), (\ref{eq:jacobian2}) it follows that:

\begin{equation}\label{eq:unrolled_jacobian_flow}
\begin{split}
\partial_Y X^{(L)} &= \partial_{X^{(L-1)}} X^{(L)} \cdots  \partial_{X^{(1)}} X^{(2)} \partial_{Y} X^{(1)}\\
&= \Big ( \prod_{l=1}^{L-1}  \mathbb{I}_{Md} - \tau \nabla_X^2 L_{C(X^{(l)}, Y)}^{\epsilon} (a, b) \Big ) \cdot \Big (- \tau \partial_Y \nabla_X L_{C(X^{(0)}, Y)}^{\epsilon} (a, b) \Big ).
\end{split}
\end{equation}
 
We consider as an example an approximated version of the Hessian, where the optimal coupling $P^*$ is fixed and does not contribute to the rederivation of $\nabla_X L_{C(X,Y)}^\epsilon(a,b)$. This is an accurate approximation for very large values of $\epsilon$, but it does not hold for the general case. We use it here however to illustrate the pathological behaviour that occurs when unrolling through the iterations of the gradient flow. We denote this Hessian as $\tilde{\nabla}^2_X L_{C(X, Y)}^{\epsilon}(a, b)$. In the case where the cost that captures the geometry of the space is that of the squared Euclidean cost $C(x_i, y_j)=(x_i - y_j)^2$, the approximated Hessian is equal to: 

\begin{equation}\label{eq:hessian_approx}
\tilde{\nabla}^2_X L_{C(X, Y)}^{\epsilon}(a, b)=\frac{2}{M} \mathbb{I}_{Md}. 
\end{equation}

By substituting Eq. (\ref{eq:hessian_approx}) in Eq. (\ref{eq:unrolled_jacobian_flow}), we obtain:

\begin{equation}\label{eq:unrolled_jacobian_flow_approx}
\begin{split}
\partial_Y X^{(L)} &= \partial_{X^{(L-1)}} X^{(L)} \cdots  \partial_{X^{(1)}} X^{(2)} \partial_{Y} X^{(1)}\\
&= \Big ( \prod_{l=1}^{L-1}  \mathbb{I}_{Md} - \tau \frac{2}{M} \mathbb{I}_{Md} \Big ) \cdot \Big (- \tau \partial_Y \nabla_X L_{C(X^{(0)}, Y)}^{\epsilon} (a, b) \Big )\\
&= \Big ( \prod_{l=1}^{L-1}  \mathbb{I}_{Md} (1 - \tau \frac{2}{M}) \Big ) \cdot \Big (- \tau \partial_Y \nabla_X L_{C(X^{(0)}, Y)}^{\epsilon} (a, b) \Big )\\
&= \Big ( \mathbb{I}_{Md} (1 - \tau \frac{2}{M})^{L-1} \Big ) \cdot \Big (- \tau \partial_Y \nabla_X L_{C(X^{(0)}, Y)}^{\epsilon} (a, b) \Big ).\\
\end{split}
\end{equation} 

We notice that when the pooled representation has $M > 2$ points it holds that $0<1-\tau \frac{2}{M}<1$. As a result, for a large number of iterations $L$ of the gradient flow, the term $(1-\tau \frac{2}{M})^{L-1}$ tends to zero and the Jacobian $\partial_Y X^{(L)}$ will also tend to zero. This can lead to the pathological behaviour of vanishing gradients \cite{pmlr-v97-metz19a}. Even though the true Hessian is not equal to the diagonal matrix in Eq. (\ref{eq:hessian_approx}), because the $P^*$ is not freezed, we observe vanishing gradients in practice. Further, when training a neural network, the automatic differentiation defaults to reverse mode and, therefore, intermediate computations of the gradient flow for all the iterations need to be stored in memory. In order to overcome these issues, we propose to backpropagate through the gradient flow using implicit differentiation, as described in the next Section.

\subsection{Implicit Differentiation of FlowPool}

\subsubsection{Implicit Function Theorem}\label{sec:implicit_diff}
The implicit function theorem \cite{krantz2012implicit} states that, given a continuously differentiable function $F$, and a point $(x^*(\theta  ), \theta)$ such that $F(x^*(\theta), \theta)=0$, if the Jacobian $\partial_1 F(x^*(\theta), \theta)$ is invertible, the variables $x(\theta)$ are differentiable functions of $\theta$ in some neighborhood of the point $(x^*(\theta), \theta)$. By applying the chain rule, we obtain:

\begin{equation}\label{eq:implicit_theorem}
\begin{split}
& \partial_1 F(x^*(\theta), \theta) \partial x^*(\theta) + \partial_2 F(x^*(\theta), \theta) = 0 \Leftrightarrow\\
& -\partial_1 F(x^*(\theta), \theta) \partial x^*(\theta) = \partial_2 F(x^*(\theta), \theta),
\end{split}
\end{equation} 

where by $\partial_1, \partial_2$ we denote the Jacobian with respect to the first and the second argument, accordingly. As a result, the Jacobian $\partial x^*(\theta)$ can be obtained by solving the linear system in Eq. (\ref{eq:implicit_theorem}). 

The recent work in \cite{blondel2021efficient} proposes to use the implicit function theorem in order to perform automatic differentiation of optimization problems. In that context, $\theta$ stands for the inputs of the optimization problem, $x^*(\theta)$ for the optimal solution and the function $F$ captures the optimality conditions. Thus, the implicit function theorem offers the possibility to obtain the Jacobian of the optimal solution with respect to the inputs of the optimization problem.

\subsubsection{Implicit Differentiation of the Wasserstein Gradient Flow}\label{sssec:implicit_flow}

We denote as $X^*(Y)$ the optimal solution at the output of FlowPool for input $Y$. By considering the continuously differentiable function F as: 

\begin{equation}
F(X(Y), Y) = \nabla_X L_{C(X(Y), Y)}^{\epsilon}(a,b)
\end{equation}

at optimality it holds that:

\begin{equation}
\begin{split}
F(X^*(Y), Y) &= \mathbb{0}_{M \times d} \Leftrightarrow \\
\nabla_X L_{C(X^{*}(Y), Y)}^{\epsilon}(a,b)&=\mathbb{0}_{M \times d}.
\end{split}
\end{equation}

Therefore, from Eq. (\ref{eq:implicit_theorem}), we can obtain the Jacobian $\partial_Y X^*(Y)$ by solving the linear system:

\begin{equation}\label{eq:implicit_diff_of_flow}
\begin{split}
-\partial_X \nabla_X L_{C(X^{*}(Y), Y)}^{\epsilon}(a,b) \partial_Y X^*(Y) &= \partial_Y \nabla_X L_{C(X^{*}(Y), Y)}^{\epsilon}(a,b) \Leftrightarrow \\
-\nabla^2_X L_{C(X^{*}(Y), Y)}^{\epsilon}(a,b) \partial_Y X^*(Y) &= \partial_Y \nabla_X L_{C(X^{*}(Y), Y)}^{\epsilon}(a,b). 
\end{split}
\end{equation}

As a result, the Jacobian $\partial_Y X^*(Y)$ of the Wasserstein gradient flow can be obtained by solving the linear system in Eq. (\ref{eq:implicit_diff_of_flow}), if the Hessian of the Sikhorn loss with respect to the pooled representation $\nabla^2_X L_{C(X(Y), Y)}^{\epsilon}(a,b)$ evaluated at the optimal solution $(X^*(Y), Y)$, is a square invertible matrix. 

In the case where the Hessian is positive definite (strictly positive eigenvalues and of full rank) the optimal solution $X^*$ of Eq. (\ref{eq:reg_flow}) is a strict local minimum. In the case where the Hessian is positive semidefinite (non-negative eigenvalues and rank deficient), $X^*$  is a candidate for a local minimum. Further, it is possible that $X^*$ is a saddle point. In that case the Hessian is indefinite and has both positive and negative eigenvalues. It should be noted that the probability that $\nabla^2_X L_{C(X^{*}(Y), Y)}^{\epsilon}(a,b)$ is indefinite is higher when instead of the Sinkhorn loss we use the Sinkhorn divergence in Eq. (\ref{eq:sinkhorn_div}) in order to remove the entropic bias. In particular, when minimizing Eq. (\ref{eq:sinkhorn_div}) with respect to $X$, only the first two terms $L_{C(X,Y)}^{\epsilon}(a,b)$ and $ - \frac{1}{2}L_{C(X,X)}^{\epsilon}(a, a)$ provide non-zero gradients with respect to $X$. Further, due to the negative sign in the autocorrelation term $L_{C(X,X)}^{\epsilon}(a, a)$, this minimization is essentially a minimization with respect to $X$ of $L_{C(X,Y)}^{\epsilon}(a,b)$ and a maximization with respect to $X$ of $L_{C(X,X)}^{\epsilon}(a, a)$. As a result, minimization of Eq. (\ref{eq:sinkhorn_div}) is likely to converge to a saddle point. Further, as the entropic regularization parameter $\epsilon$ admits larger values, the autocorrelation term $L_{C(X,X)}^{\epsilon}(a, a)$ becomes more prominent and this phenomenon is more prevalent. 

In order to account for the different spectral properties of the Hessian at the optimal solution, instead of the linear problem in Eq. (\ref{eq:implicit_diff_of_flow}), we solve the normal equation:

\begin{align}
A \, J =B,
\end{align}

where:
\begin{equation}
A=-\nabla^2_X L_{C(X^{*}(Y), Y)}^{\epsilon}(a,b)^{\top}\nabla^2_X L_{C(X^{*}(Y), Y)}^{\epsilon}(a,b)
\end{equation}

\begin{equation}
B=\nabla^2_X L_{C(X^{*}(Y), Y)}^{\epsilon}(a,b)^{\top} \partial_Y \nabla_X L_{C(X^{*}(Y), Y)}^{\epsilon}(a,b)
\end{equation}

\begin{equation}
J=\partial_Y X^*(Y).
\end{equation}

By doing so, the matrix $A$ is always positive semi-definite. Further, in order to account for the possible rank deficiency of $A$ due to zero eigenvalues, or for the ill-conditioning of $A$ in the case where its smallest singular value is close to zero, we add a Tikhonov regularization \cite{tikhonov1943stability} with parameter $\lambda$ and solve the linear system:

\begin{equation}\label{eq:linear_impl_flowpool}
(A + \lambda \mathbb{I})J =B
\end{equation}

with conjugate gradient descent \cite{stiefel1952methods} in order to retrieve the Jacobian $J=\partial_Y X^*(Y)$. 

\begin{figure}[!h]
\centering
\includegraphics[width=0.5 \linewidth]{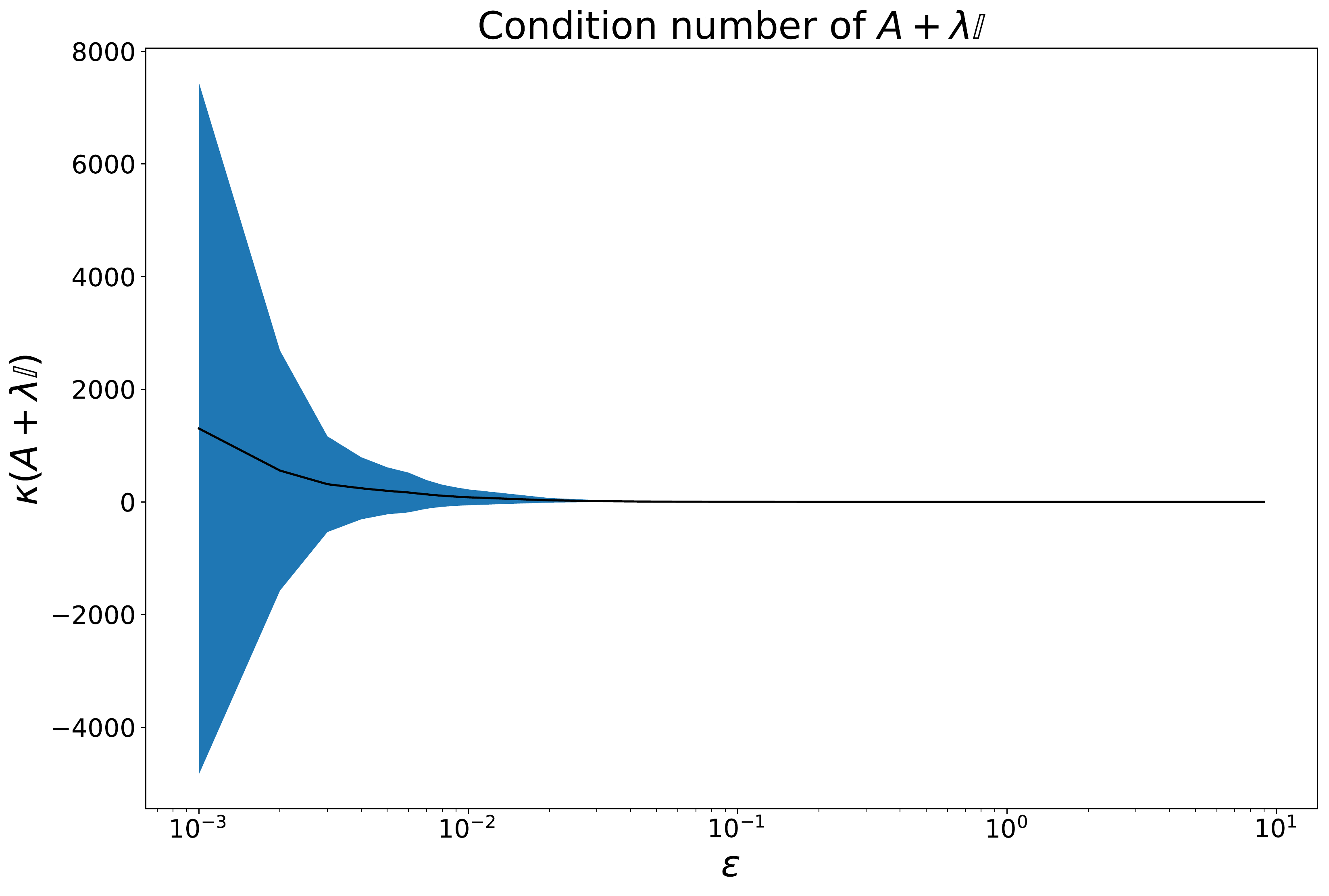}
\caption{Evolution of the condition number of $A+\lambda I$ of the linear operator in the implicit differentiation scheme of FlowPool as a function of the entropy regularization parameter $\epsilon$. The black line shows the mean and the blue shaded area the standard deviation. It can be seen that larger values of $\epsilon$ lead to better conditioning. }
\label{fig:condition_number_linear_op}
\end{figure}

In Fig. (\ref{fig:condition_number_linear_op}) we show the condition number of the linear operator $A + \lambda I$ that needs to be inverted in order to solve the system in Eq. (\ref{eq:linear_impl_flowpool}). It can be seen that for small values of $\epsilon$, the linear system is not well conditioned. As $\epsilon$ obtains larger values, the conditioning becomes better and eventually $\kappa$ tends to 1. We note that, in this case, conditioning is affected by both the entropic regularization parameter $\epsilon$ as well as the Tikhonov regularization parameter $\lambda$. 

In practice, when FlowPool is incorporated in a GNN architecture, the explicit formulation of the Jacobian $J$ is not needed. Instead, during reverse-mode automatic differentiation, it suffices to compute the vector-Jacobian products $v^{\top}J$, where $v$ corresponds to the gradient of the loss $\mathcal{L}$ used to train the neural network with respect to $X^*$. In order to obtain the vector-Jacobian product $v^{\top}J$, we follow the procedure propoced in \cite{blondel2021efficient}. First, we solve the linear system:

\begin{equation}\label{eq:linear_system_vjp}
A^{\top}u=v
\end{equation}

with respect to $u$. Then $v^{\top}J$ is obtained as:

\begin{equation}
v^{\top}J=u^{\top}AJ=u^{\top}B.
\end{equation}

This procedure allows to differentiate the output of FlowPool $X^*$ with respect to different variables by solving the linear system in Eq. (\ref{eq:linear_system_vjp}) only once. For instance, if the cost is parametrized with parameter $\theta$ as $C_\theta$, we can compute $v^\top J_\theta$ as $u^\top B_{\theta}$ where $B_{\theta}=\nabla^2_X L_{C(X^{*}(Y), Y)}^{\epsilon}(a,b)^{\top} \partial_\theta \nabla L_{C(X^*(Y),Y)}^\epsilon (a,b)$. 

\section{Experiments}\label{sec:pool_experiments}

\subsection{Illustrative Example}
\begin{figure}[h!]%
    \centering
    \subfloat[]{{\includegraphics[width=5cm]{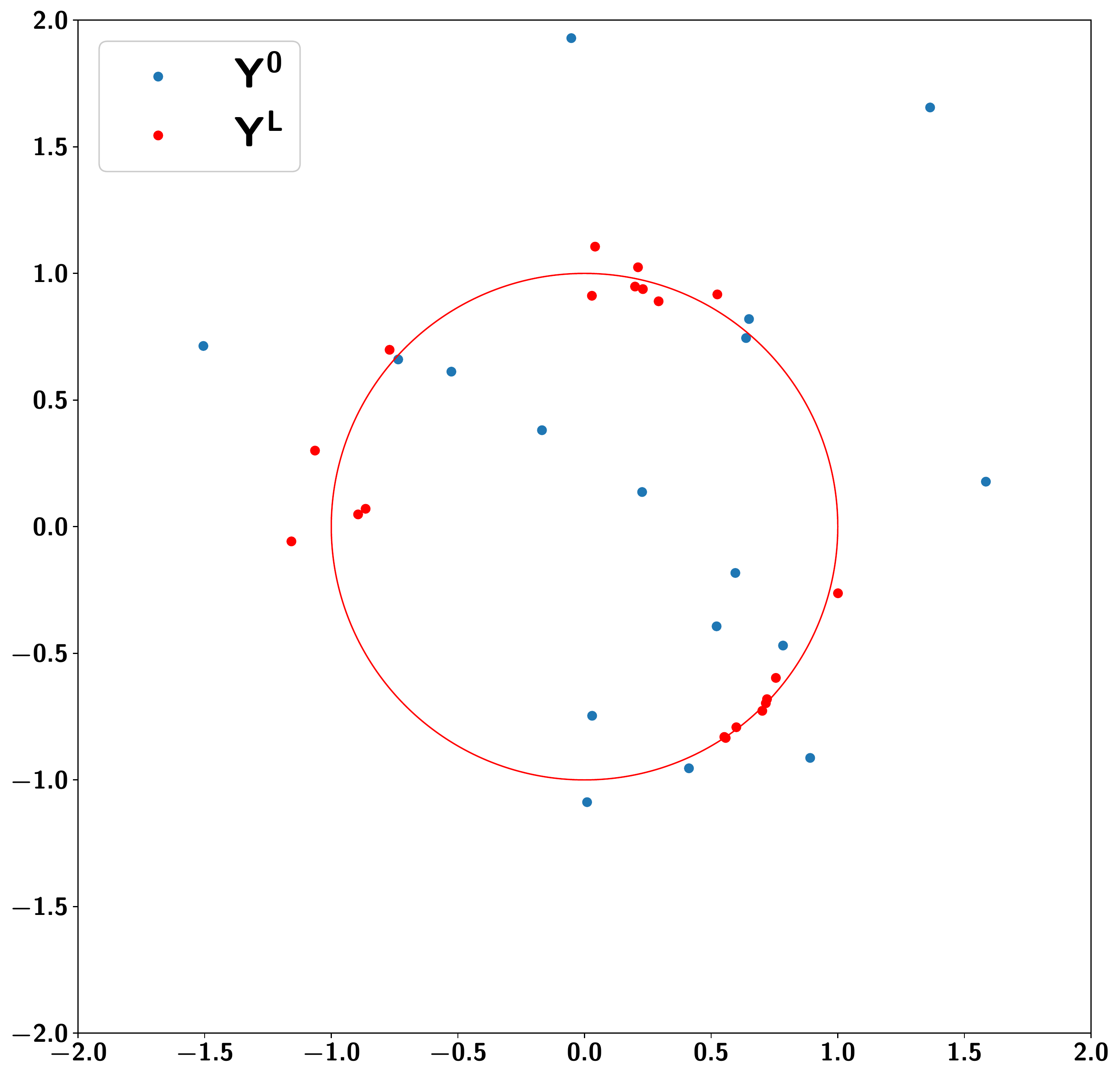} }}%
    \qquad
    \subfloat[]{{\includegraphics[width=5cm]{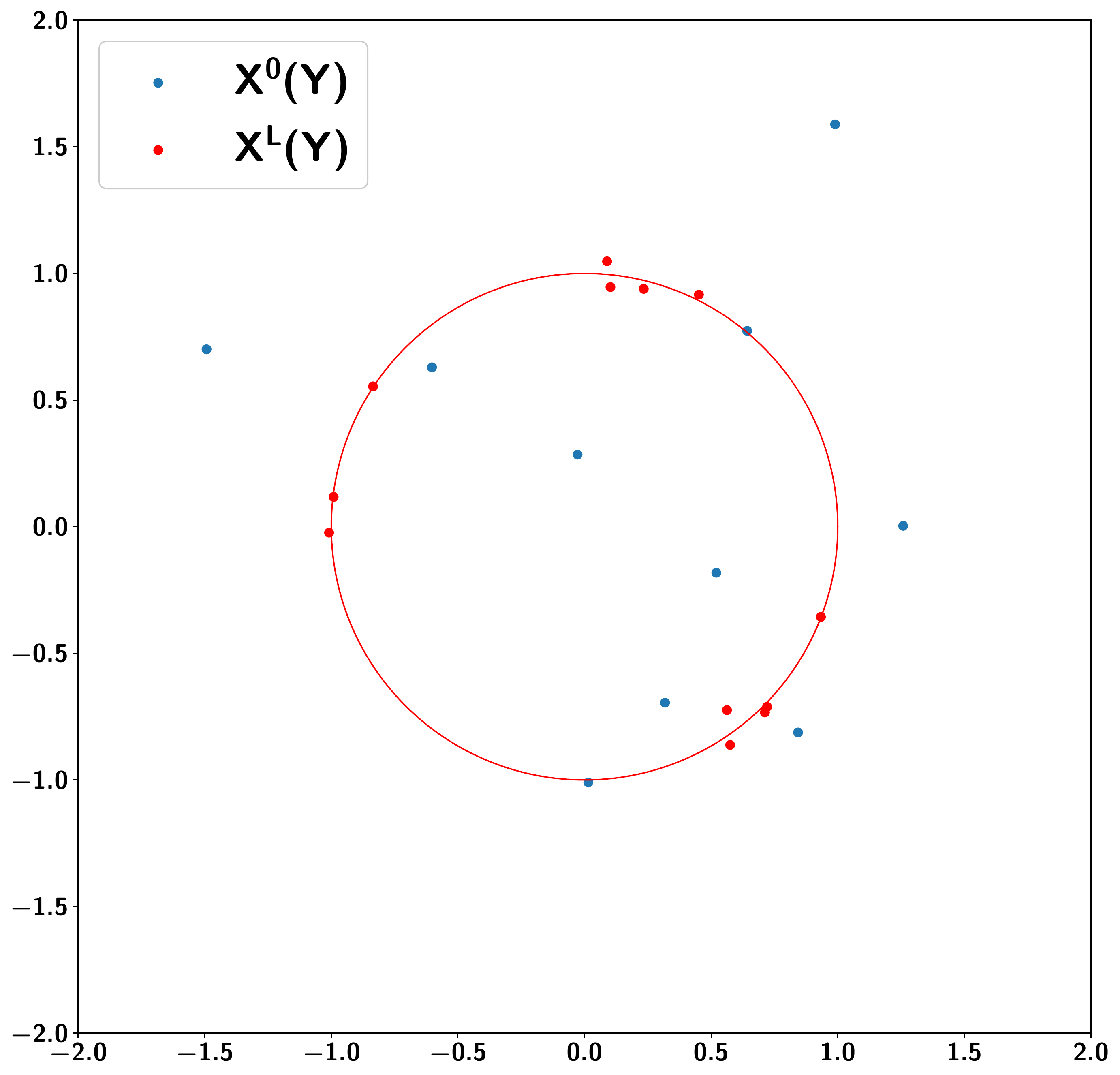} }}%
    \caption{Illustrative example. The points of the pointcloud $X$ are forced to be on the unit circle by solving the optimization problem in Eq. (\ref{eq:unit_circle_min}). It can be seen that at convergence the points of $X^L$, shown with the red points in (b), are on the unit circle. Because FlowPool minizes the Wasserstein distance between the pointclouds $X$ and $Y$, we obtain at convergence $Y^L$, whose points are also very close to the unit circle, as shown in (a).} %
    \label{fig:unit_circle}%
\end{figure}

In order to illustrate that the proposed scheme for the implicit differentiation of FlowPool works as expected, we conduct here a simple experiment. We initialize a random pointcloud $Y \in  \mathbb{R}^{20 \times 2}$ from the normal distribution, shown with blue in Fig. (\ref{fig:unit_circle}a). We input $Y$ to FlowPool with $C(x_i,y_j)=(x_i-y_j)^2$, in order to retrieve a pointcloud $X \in  \mathbb{R}^{12 \times 2}$ and solve the following minimization problem:

\begin{equation}\label{eq:unit_circle_min}
\operatorname*{min}_Y \|X(Y)^{\circ 2}1_2 - 1_{12} \|_2^2.
\end{equation}

By solving the problem in Eq. (\ref{eq:unit_circle_min}), we are forcing each point of $X$ to belong on the unit circle. It can be observed that the problem in Eq. (\ref{eq:unit_circle_min}) is differentiable with respect to $Y$ and can therefore be solved using a gradient based method. Here, we use Adam \cite{DBLP:journals/corr/KingmaB14} with a learning rate of $\mu=0.01$. In Fig. (\ref{fig:unit_circle}b) we show with red the optimal $X$ after $L$ iterations. It can be observed that all points of $X^L$ are on the unit circle. What we would expect, since we are minimizing the Wasserstein distance between the pointclouds $X$ and $Y$, is for the points of $Y^L$ to also be on the unit circle. As can be observed from Fig. (\ref{fig:unit_circle}a), the points of $Y^L$ are very close to the unit circle. Therefore, we have confirmed qualitatively that the proposed implicit differentiation scheme of FlowPool operates as expected.


\subsection{Graph Classification}
We now evaluate FlowPool on a graph classification task. We consider a GNN architecture composed of one SGCN layer \cite{wu2019simplifying} of power $K=2$, followed by a global pooling layer and a logistic regression classifier. We use for the global pooling layer either FlowPool or SortPool \cite{zhang2018end} and compare the classification performance. 

We consider the MUTAG \cite{debnath1991structure} dataset which consists of 188 samples of chemical compounds. Each chemical compound corresponds to a graph where vertices stand for atoms and edges between vertices represent bonds between the atoms. Each node is labeled by its atom type and there are 7 atom types in total. The goal is to classify the compounds according to their mutagenicity on Salmonella typhimurium. For the evaluation we consider the framework proposed in \cite{errica2020fair} for the fair comparison of GNNs in the task of
graph classification. In that framework a 10 fold cross-validation scheme is performed and the splits are stratified. For both models we use a batch size of 32. The optimizer used is Adam \cite{DBLP:journals/corr/KingmaB14} with a learning rate of $\mu=0.01$. We train over 300 epochs and perform early-stopping \cite{DBLP:conf/iclr/ZhangBHRV17} with a patience of 20 epochs. The criterion for early-stopping is based on the validation loss.

In Table \ref{tab:d8m5} we show the mean and the standard deviation of the classification accuracy obtained for $d=8$ and $M=5$ over the 10 folds. It can be seen that FlowPool significantly outperforms SortPool. With this experiment we have demonstrated the advantage of FlowPool that preserves the statistics of a graph representation to its pooled counterpart. The integration of FlowPool in more complex GNN architectures can allow its evaluation on larger datasets than the one considered in this experiment.   

\begin{table}[!h]
	\renewcommand{\arraystretch}{1}
	\centering
	\setlength\tabcolsep{0.5 pt}
	\begin{tabular}{|c|c|}
		\hline
		\textbf{Model} & \textbf{Classification Accuracy}  \\
		\hline
		SortPool & 73.30 $\pm$ 7.88 \\
		\hline
		FlowPool & 82.48 $\pm$ 7.39 \\
		\hline
	\end{tabular}
	\caption{Graph Classification on MUTAG.}
	\label{tab:d8m5}
\end{table}

\section{Parametrization of FlowPool - Learning the Ground Cost} \label{sec:parametrization}
The pooled representations returned by FlowPool can become more relevant to the specific dataset, used to train the GNN architecture, by considering a parametrization of the ground cost $C_\theta$ used for the mass transportation in order to take into account the structures of the considered graphs. This corresponds to the problem of ground metric learning and has been studied in a line of works, such as \cite{cuturi2014ground}, \cite{heitz2021ground}, \cite{wang2012supervised}, \cite{zen2014simultaneous}, among others. We believe that learning an appropriate cost from the data could offer a promising direction for future work. Our implementation of FlowPool with automatic differentiation ensures that the relevant derivatives can be computed for any cost that captures the pairwise relationship between the input representation and its pooled counterpart. Further, the implicit differentiation with respect to the parameters $\theta$ can be performed with a very small computational overhead, as already discussed in Section \ref{sssec:implicit_flow}.

\section{Conclusion}\label{sec:conclusion}
In this work we proposed FlowPool, a framework for pooling graph representations, while optimally preserving their statistical properties. Our proposed method is framed as a Wasserstein gradient flow with respect to the pooled graph representation 
We proposed a versatile implementation of our method, based on automatic differentiation, that can take into account the geometry of the representation space through any optimal transport cost. Further, we proposed an implicit differentiation scheme for the FlowPool layer in order to overcome the issue of vanishing gradients and reduce significantly memory requirements. Finally, we evaluated the performance of the proposed layer, when integrated in a GNN, on a graph classification task.

\bibliographystyle{IEEEtran}
\bibliography{bibliography}


\end{document}